\documentclass[10pt,twocolumn,letterpaper]{article}

\usepackage{iccv}
\usepackage{times}
\usepackage{epsfig}
\usepackage{graphicx}
\usepackage{amsmath}
\usepackage{amssymb}

\usepackage{booktabs}       
\usepackage{multirow}
\usepackage{makecell}
\usepackage{amsfonts}       
\usepackage{nicefrac}       
\usepackage{xcolor}         
\usepackage{algorithm}
\usepackage{algorithmic}
\usepackage{url}
\usepackage[accsupp]{axessibility}
\usepackage[pagebackref=true,breaklinks=true,colorlinks,bookmarks=false]{hyperref}
\usepackage[capitalize]{cleveref}
\crefname{section}{Sec.}{Secs.}
\Crefname{section}{Section}{Sections}
\Crefname{table}{Table}{Tables}
\crefname{table}{Tab.}{Tabs.}

\iccvfinalcopy 

\def\iccvPaperID{6322} 

\ificcvfinal\fi

\begin{document}

\title{DELFlow: Dense Efficient Learning of Scene Flow for Large-Scale Point Clouds}

\author{Chensheng~Peng\textsuperscript{\rm 1}, Guangming~Wang\textsuperscript{\rm 1}, Xian~Wan~Lo\textsuperscript{\rm 1}, Xinrui~Wu\textsuperscript{\rm 1}, Chenfeng~Xu\textsuperscript{\rm 2}, \\
Masayoshi~Tomizuka\textsuperscript{\rm 2},
Wei~Zhan\textsuperscript{\rm 2}, and Hesheng~Wang\textsuperscript{\rm 1}\thanks{ Corresponding Authors. The first two authors contributed equally.}\\
{\textsuperscript{\rm 1}Department of Automation, Key Laboratory of System Control and Information Processing of}\\
{Ministry of Education, Shanghai Jiao Tong University}\\
{\textsuperscript{\rm 2} Mechanical Systems Control Laboratory, University of California, Berkeley} \\
\small{\texttt{\{pesiter-swift,wangguangming,kaylex.lo,916806487,wanghesheng\}@sjtu.edu.cn}} \\
\small{\texttt{\{xuchenfeng,tomizuka,wzhan\}@berkeley.edu}
}
}
%

\maketitle
\ificcvfinal\thispagestyle{empty}\fi

\begin{abstract}
Point clouds are naturally sparse, while image pixels are dense. The inconsistency limits feature fusion from both modalities for point-wise scene flow estimation. Previous methods rarely predict scene flow from the entire point clouds of the scene with one-time inference due to the memory inefficiency and heavy overhead from distance calculation and sorting involved in commonly used farthest point sampling, KNN, and ball query algorithms for local feature aggregation. To mitigate these issues in scene flow learning, we regularize raw points to a dense format by storing 3D coordinates in 2D grids. Unlike the sampling operation commonly used in existing works, the dense 2D representation 1) preserves most points in the given scene, 2) brings in a significant boost of efficiency, and 3) eliminates the density gap between points and pixels, allowing us to perform effective feature fusion. We also present a novel warping projection technique to alleviate the information loss problem resulting from the fact that multiple points could be mapped into one grid during projection when computing cost volume. Sufficient experiments demonstrate the efficiency and effectiveness of our method, outperforming the prior-arts on the FlyingThings3D and KITTI dataset. 
Our source codes will be released on \url{https://github.com/IRMVLab/DELFlow}.

\end{abstract}

\section{Introduction}
\label{sec:intro}

Scene flow represents the point-wise motion between a pair of frames, specifically the magnitude and direction of the 3D motion. As a low-level task, 3D scene flow estimation is beneficial to numerous high-level scene understanding tasks in autonomous driving, such as LiDAR odometry \cite{wang2021pwclo}, object tracking \cite{wang2020pointtracknet}, and semantic segmentation \cite{liu2019meteornet}.

\begin{figure}[t]
	\centering
	\includegraphics[width=0.98\linewidth]{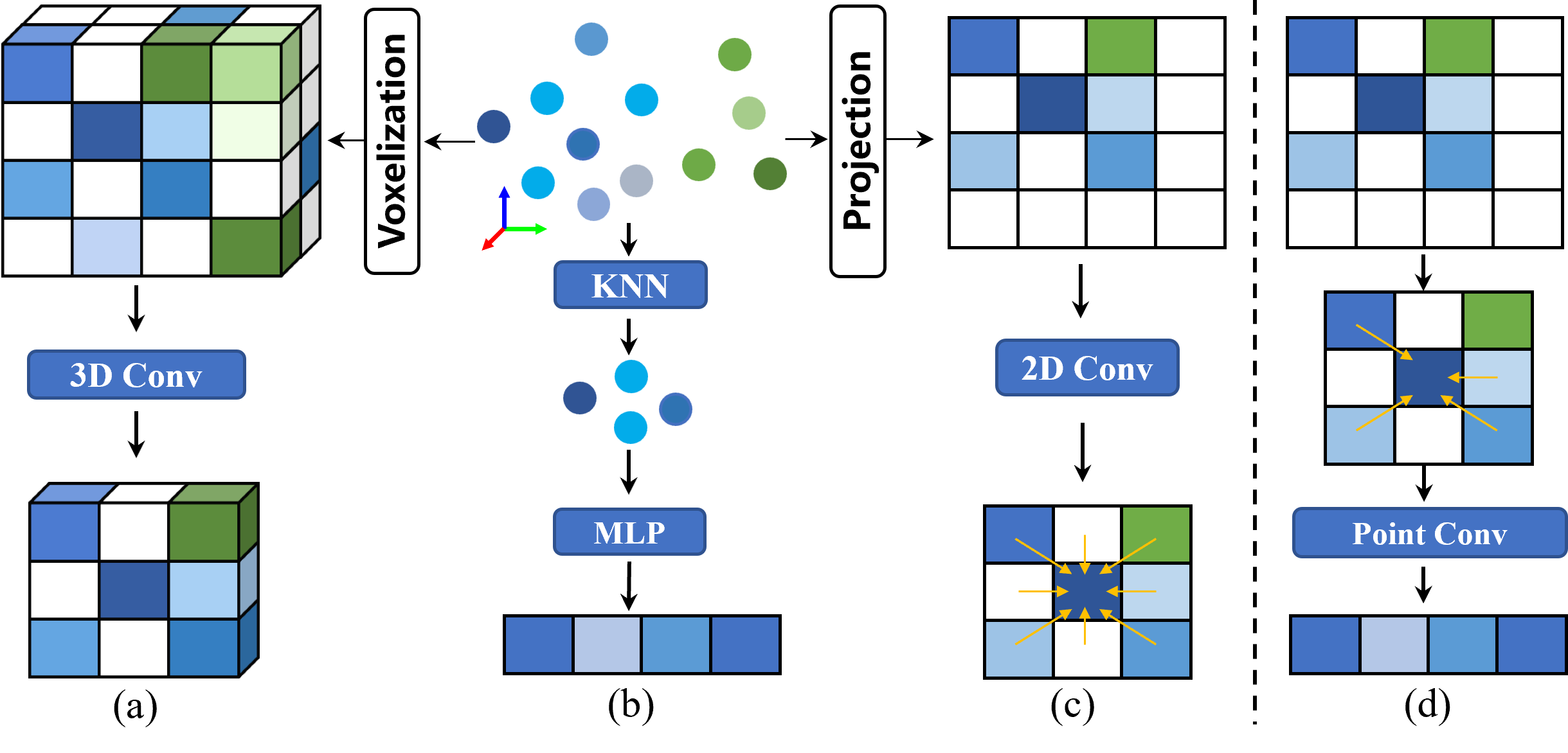}
    \vspace{-1mm}
	\caption{Comparison of current point cloud processing frameworks. (a) 3D grid-based methods. (b) 3D point-based methods. (c) 2D perspective grid-based methods. (d) Our outlier-aware point-based methods, where outliers are filtered out.}
	\vspace{-4mm}
	\label{fig:represent}
\end{figure}

Early works \cite{choy20163d, riegler2017octnet} convert point cloud into 3D voxel grids and process them with 3D Convolutional Neural Networks (CNNs). However, the computational costs of such voxel-based methods grow drastically with the resolution. Besides voxelization (Fig. \ref{fig:represent}\textcolor{red}{a}), recent studies \cite{gu2019hplflownet,liu2019flownet3d,wu2019pointpwc, wang2021unsupervised} resort to learning scene flow directly from the raw coordinates of point clouds in 3D space. These point-based methods \cite{liu2019flownet3d, wu2019pointpwc, wang2021hierarchical, wang2022matters} (Fig. \ref{fig:represent}\textcolor{red}{b}) relies on feature aggregation from neighborhood points with a PointNet \cite{qi2017pointnet, qi2017pointnet++} structure. Nevertheless, the most commonly used \textit{K} Nearest Neighbors (KNN) and ball query algorithms in solutions \cite{gu2019hplflownet,liu2019flownet3d,wu2019pointpwc, wang2022residual} require frequent distance calculation and sorting between point clouds, which are memory-inefficient and time-consuming. Therefore, only a limited number of points are taken as inputs because large-scale input point clouds indicate more GPU memory consumption. The inefficiency led to the proposal of projection-based methods \cite{boulch2017unstructured, wang2018fusing,wu2018squeezeseg}, where 2D convolutions are applied to the projected point cloud on image plane. As shown in Fig. \ref{fig:represent}\textcolor{red}{c}, a convolution kernel is used to perform feature extraction among nine neighbors. Such 2D perspective grid-based methods suffer from the neighboring outliers and are not flexible since effective 3D methods cannot be applied.

To deal with the heavy computational cost of 3D methods and the limited accuracy of 2D methods, we propose a projection-based framework for scene flow learning from dense point clouds. We store raw 3D point clouds ($n \times 3$) in the corresponding 2D pixels by projecting them onto image plane ($H \times W \times 3$). Such representation allows us to take the complete point clouds from input scene. For the second stage, we propose \textit{Kernel Based Grouping} to capture 3D geometric information on 2D grid. The operations are conducted on 2D grids, but the output is in the 3D space. Different from 3D point-based methods, we are able to process point clouds with less memory consumption as more prior spatial information from the 2D format reduces the local query complexity. By reducing the grouping range in search of neighboring points, our
proposed network achieves single-frame feature extraction and inter-frame correlation for tens of thousands of LiDAR point clouds. In addition, the far-away points can be removed using our outlier-aware method (Fig. \ref{fig:represent}\textcolor{red}{d}), solving the problem of 2D methods caused by outliers.

The second challenge pertains to the feature fusion between point clouds and images, as raw point clouds in 3D space are naturally sparse, unordered and unstructured, while pixels of images are dense and adjacent to each other. CamLiFlow \cite{liu2022camliflow} projects a small number of sparse points onto image plane to fuse features from these two modalities. Instead, we take all points as input using the dense format of projected point clouds, which eliminates the density gap between sparse points and dense pixels, enabling effective feature fusion. As a result, we explore the potential of attentive feature fusion between dense points and pixels for more accurate scene flow prediction. 

In addition, we adopt a warping operation in cost volume to refine the predicted flow. During the refinement process, multiple warped points can be projected into the same grid on 2D plane. Previous methods merge extra points, but cause information loss which affects the final accuracy. On the contrary, our proposed cost volume module equipped with a novel warping projection technique enables us to avoid such information loss  by using the warped coordinates as an intermediate indexing variable. 


Overall, our contributions are as follows:
\vspace{-0.1cm}
\begin{itemize}
    \vspace{-0.2cm}
    \item We propose an efficient scene flow learning framework operated on projected point clouds to process points in 3D space, which takes the entire point clouds as input and predicts flow with one-time inference.
    \vspace{-0.3cm}
    \item We present a novel cost volume module with a warping projection technique, allowing us to compute the cost volume without information loss caused by merging points during the refinement process.
    \vspace{-0.3cm}
    \item We design a pixel-point feature fusion module, which encodes color information from images to guide the decoding of point-wise motion in point clouds, improving the accuracy of scene flow estimation.
\end{itemize}

\section{Related work}
\noindent{}{\bf Scene Flow Learning from 3D Data.}
With the development of deep learning for raw 3D point cloud processing \cite{qi2017pointnet,qi2017pointnet++}, FlowNet3D \cite{liu2019flownet3d} pioneers in directly processing point clouds and predicts 3D scene flow in an end-to-end fashion. A flow embedding layer is proposed to compute the correlation between a pair of point clouds. PointPWC-Net \cite{wu2019pointpwc} proposes a patch-to-patch method by considering more than just one point of the first frame during the correlation process, and extends the coarse-to-fine structure from optical flow \cite{sun2018pwc} to scene flow estimation. Inspired by permutohedral lattice \cite{adams2010fast} and Bilateral Convolutional Layers (BCL) \cite{jampani2016learning}, HPLFlowNet \cite{gu2019hplflownet} proposes DownBCL, UpBCL, and CorrBCL to restore rich information from point clouds. However, the interpolation from points to permutohedral lattice leads to information loss. To assign different weights to the correlated points in a patch, HALFLOW \cite{wang2021hierarchical} proposes a hierarchical neural network with a double attentive embedding layer. Using optimal transport \cite{petric2019got, peyre2016gromov, titouan2019optimal} tools, FLOT \cite{puy2020flot} achieves competitive performance with much less parameters. The rigidity assumption is widely used in other works \cite{vogel2013piecewise, vogel20153d, menze2015object, ma2019deep}. HCRF-Flow \cite{li2021hcrf} formulates the rigidity constraint as a high order term. Built on the architecture of RAFT \cite{teed2020raft}, RAFT-3D \cite{teed2021raft} utilizes rigid-motion embeddings to group neighbors into rigid objects and refines the 2D flow field with a recurrent structure. 


\noindent{}{\bf Efficient Learning of Point Clouds.}
Previous works \cite{choy20163d, riegler2017octnet} convert point clouds into voxel grids and process them with 3D CNNs. The information loss happens during voxelization because multiple points can be mapped into one grid, and only one point is kept \cite{tang2020searching}. Such voxel-based methods are not memory-efficient due to high computational requirements with increasing voxel resolution \cite{liu2019point}. Therefore, projection-based methods \cite{boulch2017unstructured, wang2018fusing,wu2018squeezeseg, wu2019squeezesegv2, xu2020squeezesegv3} are proposed to tackle this problem. The 3D point clouds are first projected onto a 2D plane, and then 2D CNNs are applied. DarkNet53 \cite{behley2019semantickitti} applies a fully convolutional neural network on the projected points after spherical projection. However, projection-based methods usually come with loss of geometry information \cite{tang2020searching}. 
Recent point-based methods operate on raw point clouds without a voxelization process. For example, PointNet \cite{qi2017pointnet} uses Multi-Layer Perception (MLP) to extract global features directly from raw point clouds. PointNet++ \cite{qi2017pointnet++} explored the ability to learn local features from neighborhood points. Methods \cite{liu2019flownet3d, wang2021hierarchical, wu2019pointpwc} based on PointNet++ are efficient and effective for small-scale point clouds, but the computational cost increases greatly when it comes to large-scale points. YOGO \cite{xu2021you} accelerates KNN and ball query algorithms by dividing points into several sub-regions and operating on the tokens extracted from sub-regions. 
Wang \textit{et al}. \cite{wang2021efficient} proposes a cylindrical projection method in LiDAR odometry. 


\begin{figure*}[t]
	\centering
	\includegraphics[width= 0.95\textwidth]{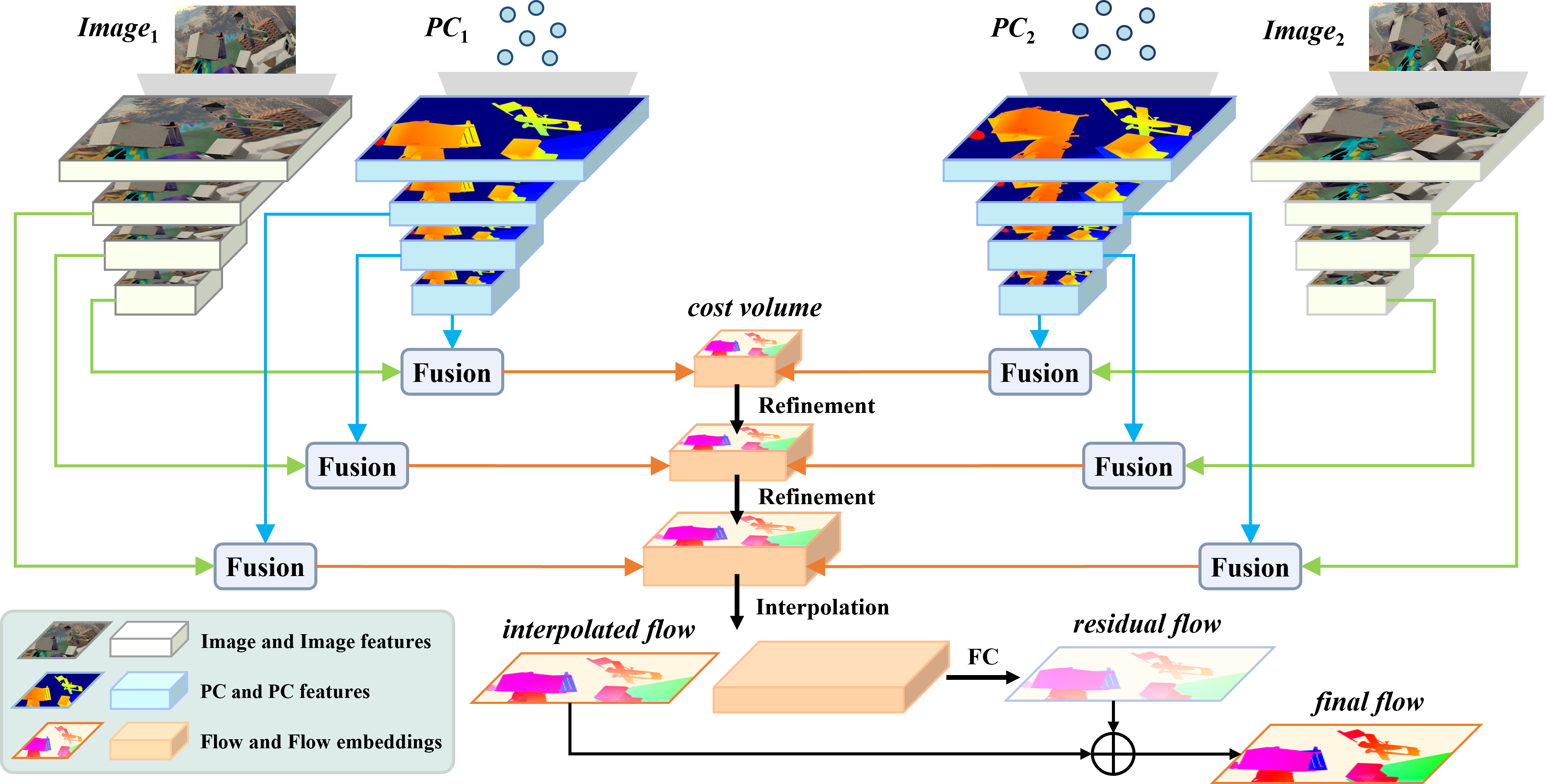}
	\caption{Structure overview of our proposed network. Given a pair of consecutive point clouds and images, down-sampling, correlation, and up-sampling is performed in order. The network predicts flow in a coarse-to-fine manner with the refinement module. Feature fusion between image and point cloud is performed before the refinement process. We adopt the residual scene flow prediction manner, by predicting a residual flow from the flow embeddings, and then adding the residual flow to the coarse flow, generating a refined flow.}
	\label{fig:sto}
\end{figure*}

\section{Efficient Learning of Dense Point Clouds}
\label{headings}

Let $PC_1 = \{ {x_i}|{x_i} \in {\mathbb{R}^3}\} _{i = 1}^{N_1}$ and $PC_2 = \{ {y_j}|{y_j} \in {\mathbb{R}^3}\} _{j = 1}^{N_2}$ represent the point clouds from a pair of consecutive frames, the goal of scene flow estimation is to predict the motion $F = \{ {\Delta x_i}|{\Delta x_i} \in {\mathbb{R}^3}\} _{i = 1}^{N_1} $ of every point in $PC_1$. Local feature aggregation from neighboring points is widely used to extract high-level point features. To achieve efficient learning of dense point clouds, we introduce a dense  format of point clouds, which brings in more prior spatial information and reduces computational complexity. Moreover, an efficient network with a new cost volume module is constructed to further improve the performance. 


\subsection{2D Representation of Dense Point Clouds}
\label{sec:representation}

\begin{figure}[t]
	\centering
	\includegraphics[width= \linewidth]{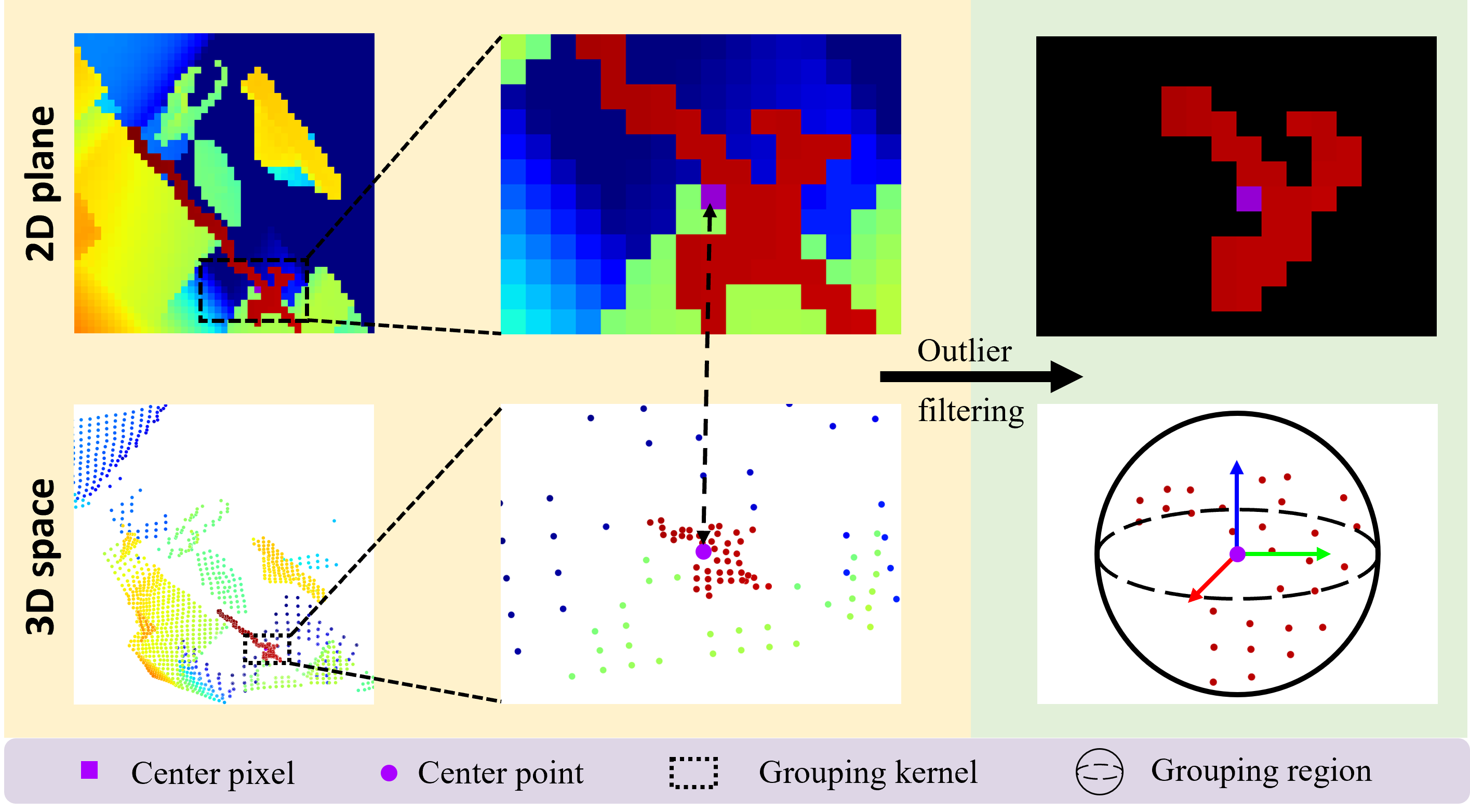}
    \vspace{-3mm}
	\caption{Kernel Based Grouping. Purple pixel denotes the central point $x_{i}^{'}$, and the black rectangle is the given kernel. Red pixels in the third column are the selected neighbors and far-away outliers are removed during the filtering process.}
\vspace{-4mm}\label{fig:intra}
\end{figure}

The original 3D point clouds ($n \times 3$) are unordered, where $n$ is the number of points. Such representation provides no prior knowledge and requires much effort to structure the irregular data \cite{tang2020searching}. To avoid unnecessary calculations, a representation of point clouds with more prior spatial information is essential for efficient processing of large-scale point clouds. Inspired by a cylindrical projection technique from \cite{wang2021efficient}, we introduce a dense representation of 3D point clouds to scene flow prediction. We conduct pixelization by projecting the raw point clouds $(x,y,z)$ onto the 2D plane $(u,v)$ with a calibration matrix.

Different from previous projection-based methods which contain RGB and depth information in the pixel grids, we fill $(u,v)$ with the 3D coordinates $(x,y,z)$ of each point, generating a $H \times W \times 3$ matrix, where $H$ and $W$ are height and width of the projected image plane. Not every pixel has its corresponding point in 3D space, therefore, we fill empty pixels with $(0, 0, 0)$. Our pixelization technique enables us to preserve complete information from raw point clouds compared to the interpolation in \cite{gu2019hplflownet} and sampling operation in \cite{liu2019flownet3d, wu2019pointpwc, wang2021hierarchical}. It allows us to process the entire LiDAR scans of the scene at once. Our scene flow prediction network takes the entire point clouds as input like an image, but with 3D coordinates instead of RGB value in the corresponding pixels. 

Instead of applying 2D solutions on the projected point clouds as other projection-based methods \cite{chai2021point, fan2021rangedet}, our approach addresses the scene flow prediction problem in a 3D manner, yet with operations on the 2D plane, ensuring the efficiency of 2D methods and effectiveness of 3D methods.

\subsection{Scene Flow Prediction Network}

Following the projection technique in Section \ref{sec:representation}, each point's coordinate is stored in the corresponding projected pixel position, and all points of the scene are taken as input. Since the resolution of point cloud is currently much higher than that of randomly sampling a small number of points, an efficient algorithm to reduce the computational cost is in need. Therefore, we construct an efficient 3D scene flow estimation network based on the 2D representation form as shown in Fig. \ref{fig:sto}. We adopt a hierarchical structure to predict the scene flow in a coarse-to-fine manner. 

\vspace{5pt}
\noindent{}{\bf Kernel Based Neighbors Grouping:}
Upon the 2D map of 3D point clouds, a kernel based grouping technique is adopted from \cite{wang2021efficient} to select neighboring points. In most cases, two nearby points in 3D space are also close on 2D plane after projection, which helps filter out many inappropriate neighborhood points without extra calculations.

We implement a CUDA operator to search for neighboring points in 3D space around the center on 2D plane. As shown in Fig. \ref{fig:intra}, given the central point $p_c$ with corresponding 2D coordinate $(u_c, v_c)$, a 2D kernel is placed around the center. We can easily locate its neighboring points within the kernel on 2D plane. The kernel size for grouping is denoted as $k_s = [k_h \times k_w]$, so we only need to search the points within range: $[u_c-k_w / 2:u_c+k_w / 2, v-k_h/2:v+k_h/2]$ instead of the whole point clouds. This technique enables us to reduce the searching complexity from $\mathcal{O}(n^2)$ to $\mathcal{O}(n\cdot k_s)$. Unlike traditional 2D kernels that operate on projected points directly, our approach allows us to remove far-away outliers that could be harmful to feature extraction. Then, the features of valid neighboring pixels are aggregated to the center.  The grouping and filtering operations are conducted on 2D plane, but the feature extraction process happens in 3D space. As a result, our proposed method shares the efficiency of 2D methods and accuracy of 3D methods. It is noteworthy that larger kernel indicates more searching time and reducing the kernel size results in a lower latency. However, if the kernel size is too small, the algorithm cannot find enough appropriate neighboring points. Through experiments, we find that an appropriate kernel size is usually between $3K \sim 4K$ where $K$ is the number of neighbors to select.

\vspace{5pt}
\noindent{}{\bf Point Cloud Encoder:}
The input point clouds are stored in a form of image, with the size of $h \times w \times 3$. For high-level feature learning, we attempt to aggregate local features from neighboring points to the central point using MLP. 

Given the raw point sets $P = \{ {x_i}|{x_i} \in {\mathbb{R}^3}\} _{i = 1}^{h \times w}$, we first select a number of central points $P_{c} = \{ {x_{i}^{'}}|{x_{i}^{'}} \in {\mathbb{R}^3}\} _{i = 1}^{h^{'} \times w^{'}}$ by setting a fixed stride. The corresponding features of $P$ and $P_{c}$ are denoted as $\{ {f_i}|{f_i} \in {\mathbb{R}^C}\} _{i = 1}^{h \times w}$ and $\{ {f_{i}^{'}}|{f_{i}^{'}} \in {\mathbb{R}^C}\} _{i = 1}^{{h^{'} \times w^{'}}}$. We use kernel based grouping method to select $K$ neighbors for each point $x_{i}^{'}$. Then, MLP is used to extract the neighboring points' features $\{ {(x_{ik}^{'}, f_{ik}^{'})}|{x_{ik}^{'}} \in {\mathbb{R}^3}, {f_{ik}^{'}} \in {\mathbb{R}^C}\} _{k = 1}^K$, generating high-level features:
 \vspace{-4pt}
\begin{equation}
\label{eq:down}
 h_{i}^{'} = \underset{k}{\textit{MaxPool}}\ \{\mathbf{MLP}((x_{ik}^{'} - x_{i}^{'}) \oplus f_{ik}^{'})\}.
 \vspace{-2mm}
\end{equation}

\begin{figure}[t]
	\centering
	\includegraphics[width=\linewidth]{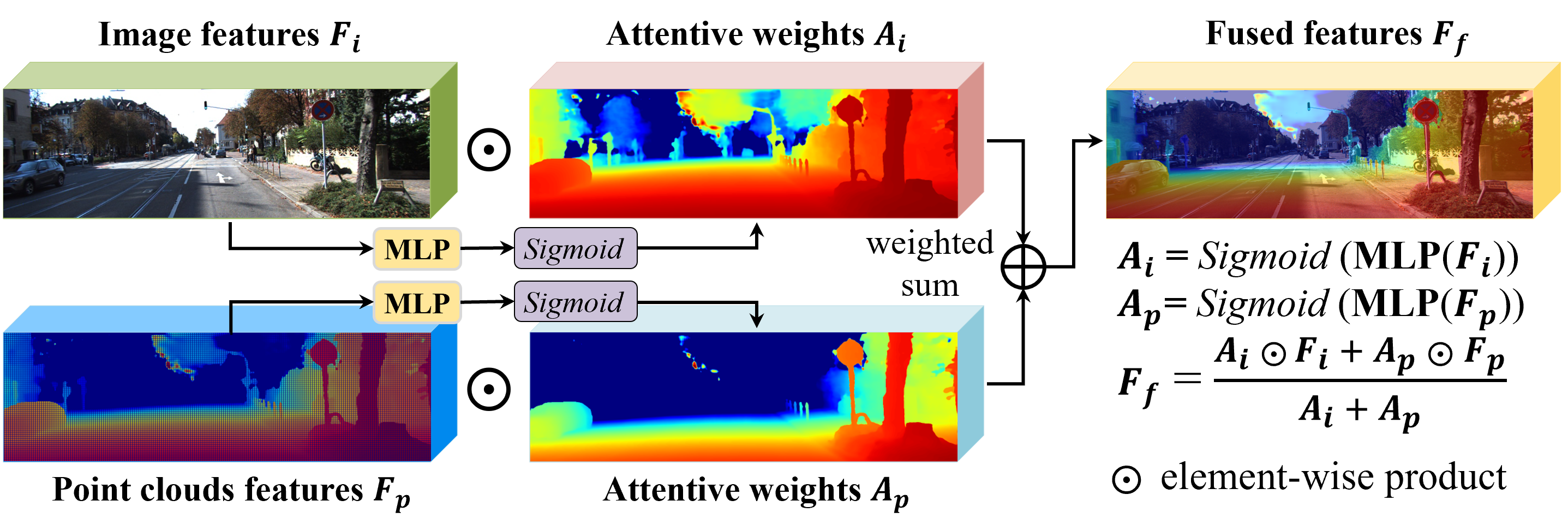}
    \vspace{-3mm}
	\caption{Feature fusion between image and point cloud features. We utilize a self-attention mechanism to align 2D and 3D features.}
	\vspace{-3mm}
	\label{fig:fusion}
\end{figure}

\noindent{}{\bf Multi-modal Feature Fusion:} 
Given the image features $ F_{i} \in {\mathbb{R}^{H \times W \times C_1}}$  and point cloud features $ F_{p} \in {\mathbb{R}^{H \times W \times C_2}}$, we utilize a self-attention mechanism to perform feature fusion between two modalities to improve the accuracy of scene flow prediction as shown in Fig. \ref{fig:fusion}. 
Since the point cloud and image share the same size of $H \times W$, each image pixel corresponds to a point and vice versa. Therefore, it is straightforward to integrate color information from 2D images to guide the prediction of 3D point-wise motion with the feature fusion module.

\vspace{5pt}
\noindent{}{\bf Attentive Feature Correlation of Point Clouds:}
After obtaining the high-level features of $PC_1$ and $PC_2$, we need to compute their motion correlation. We use the attentive cost volume in \cite{wang2021hierarchical} to learn the hidden motion pattern between two consecutive frames of point clouds. 

Given point clouds $PC_1 = \{ {(x_i, f_i)}|{x_i} \in {\mathbb{R}^3}, {f_i} \in {\mathbb{R}^C} \} _{i = 1}^{N_1}$ and $PC_2 = \{ {(y_j, g_j)}|{y_j} \in {\mathbb{R}^3}, {g_j} \in {\mathbb{R}^C}\} _{j = 1}^{N_2}$, the features of neighboring $PC_2$ are first aggregated to the centering $PC_1$ via attention mechanism and each neighbor is weighed differently. In the second stage, the features of aggregated $PC_1$ will be updated again by aggregating the neighboring $PC_1$, generating the attentive flow embedding features $ E = \{ {(x_i, e_i)}|{x_i} \in {\mathbb{R}^3}, {e_i} \in {\mathbb{R}^{C^{'}}} \} _{i = 1}^{N_1}$. We re-implement the double attentive flow embedding layers under 2D representation to accelerate the correlation between two frames of projected point clouds. 


\begin{figure}[t]
	\centering
	\includegraphics[width= \linewidth]{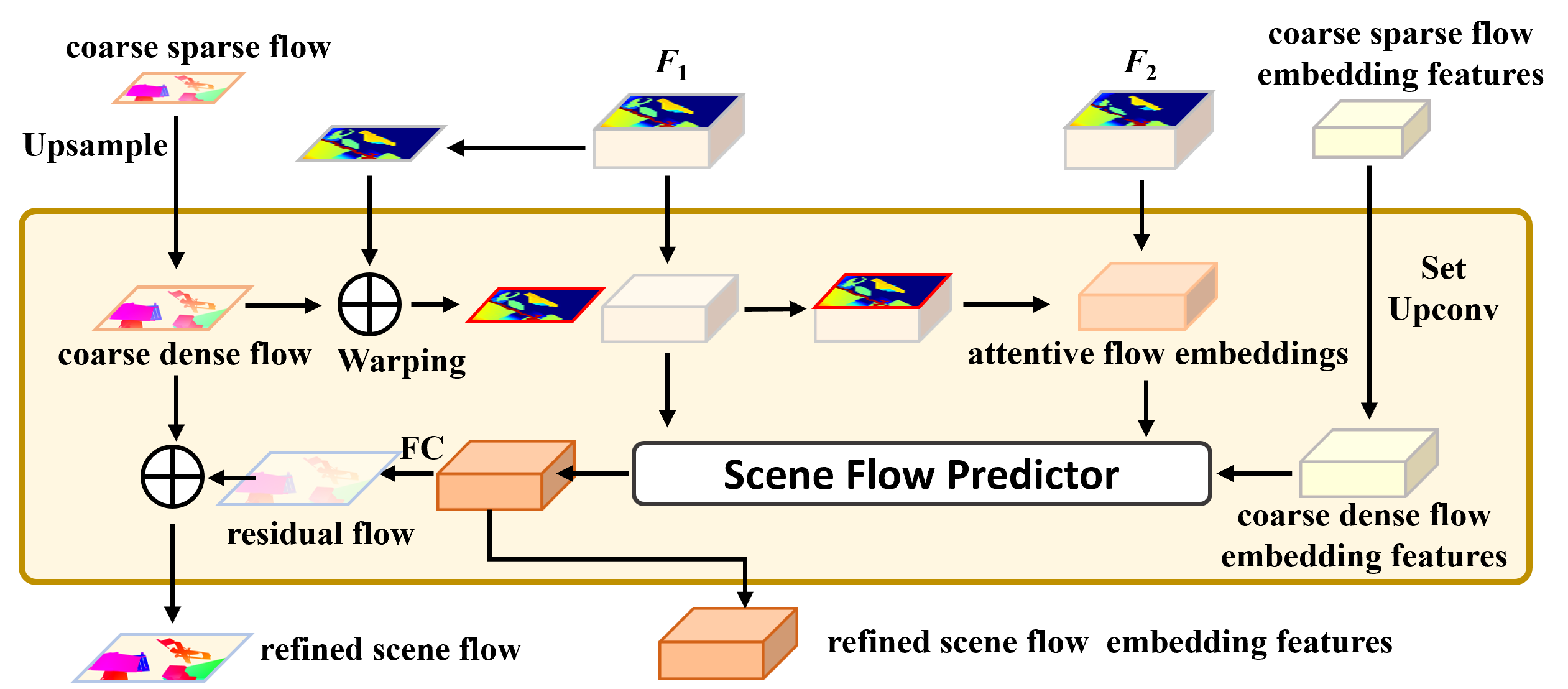}
	\vspace{-3mm}
	\caption{Residual refinement module. Set Upconv denotes up-sampling layers with learnable parameters.}
	\vspace{-4mm}
	\label{fig:refinement}
\end{figure}

\vspace{5pt}
\noindent{}{\bf Scene Flow Predictor:} The scene flow predictor takes three inputs: warped $PC_1$ features, attentive flow embedding features and the upsampled coarse dense flow embedding features. Following \cite{wu2019pointpwc, wang2021hierarchical, gu2019hplflownet}, these three inputs are concatenated together and then processed by a shared MLP, generating the refined flow embedding features.

\vspace{5pt}
\noindent{}{\bf Hierarchical Flow Refinement:}
\label{sec:refine}
We adopt the pyramid structure for supervised learning of scene flow estimation. Both the low-level and high-level scene flow is used to compute the training loss. We first predict a flow with the least number of points from the highest-level features. Then level by level, scene flow with higher resolution is predicted through up-sampling and refinement modules. For up-sampling, similar operations in equation (\ref{eq:down}) are performed among low-resolution points for high-resolution points to obtain neighboring features $b_{i}^{'}$, which are then concatenated with the fused features $s_{i}^{'}$ of central points. A shared MLP is used to generate up-sampled features.

After up-sampling operation, we obtain a coarse flow embeddings of higher-resolution points. Next, a warping operation in Section \ref{sec:costvolume} is adopted to generate the flow re-embedding features. As shown in Fig. \ref{fig:refinement}, residual scene flow can be predicted by the scene flow predictor.  Finally, we obtain a refined flow of higher accuracy by adding the estimated residual flow to the coarse flow. 


\subsection{A New Cost Volume with Warping Projection}
\label{sec:costvolume}


Assuming that $p \in \mathbb{R}^3 $ of $PC_1$ flows to $q \in \mathbb{R}^3 $ of $PC_2$, their corresponding 2D coordinates are $(u_1, v_1)$ and $(u_2, v_2)$. When there is a large motion of $p$ across frames, $(u_1, v_1)$ is far away from $(u_2, v_2)$. This requires a large searching kernel because we aim to correctly locate the corresponding $(u_2, v_2)$ in the bounding kernel around $(u_1, v_1)$. However, increasing kernel size leads to longer searching time and heavier computational cost. 

To address the issue, we incorporate the warping idea by adding the predicted coarse flow $F_{c}$ to $PC_1$, which generates $PC_{1w}=\{ {\hat{p}}|{\hat{p}} \in {\mathbb{R}^{3}}\} _{i = 1}^{N}$. The warped points $PC_{1w}$ are closer to $PC_2$ than the original $PC_1$. Next, the cost volume is computed between $PC_{1w}$ and $PC_2$ to predict a residual flow $F_{\Delta}$. By iteratively refining the predicted flow, the warped points become progressively closer to $PC_2$. We can obtain a more accurate flow $F_{a}$ by adding the residual flow to the coarse flow level by level.  
 \vspace{-4pt}
\begin{equation}
\label{eq:refine}
 F_{a} = F_{c} + \sum_{i=1}^{l} F_{\Delta i}
 \vspace{-2mm}
\end{equation}

 \begin{figure}[t]
	\centering
	\includegraphics[width=\linewidth]{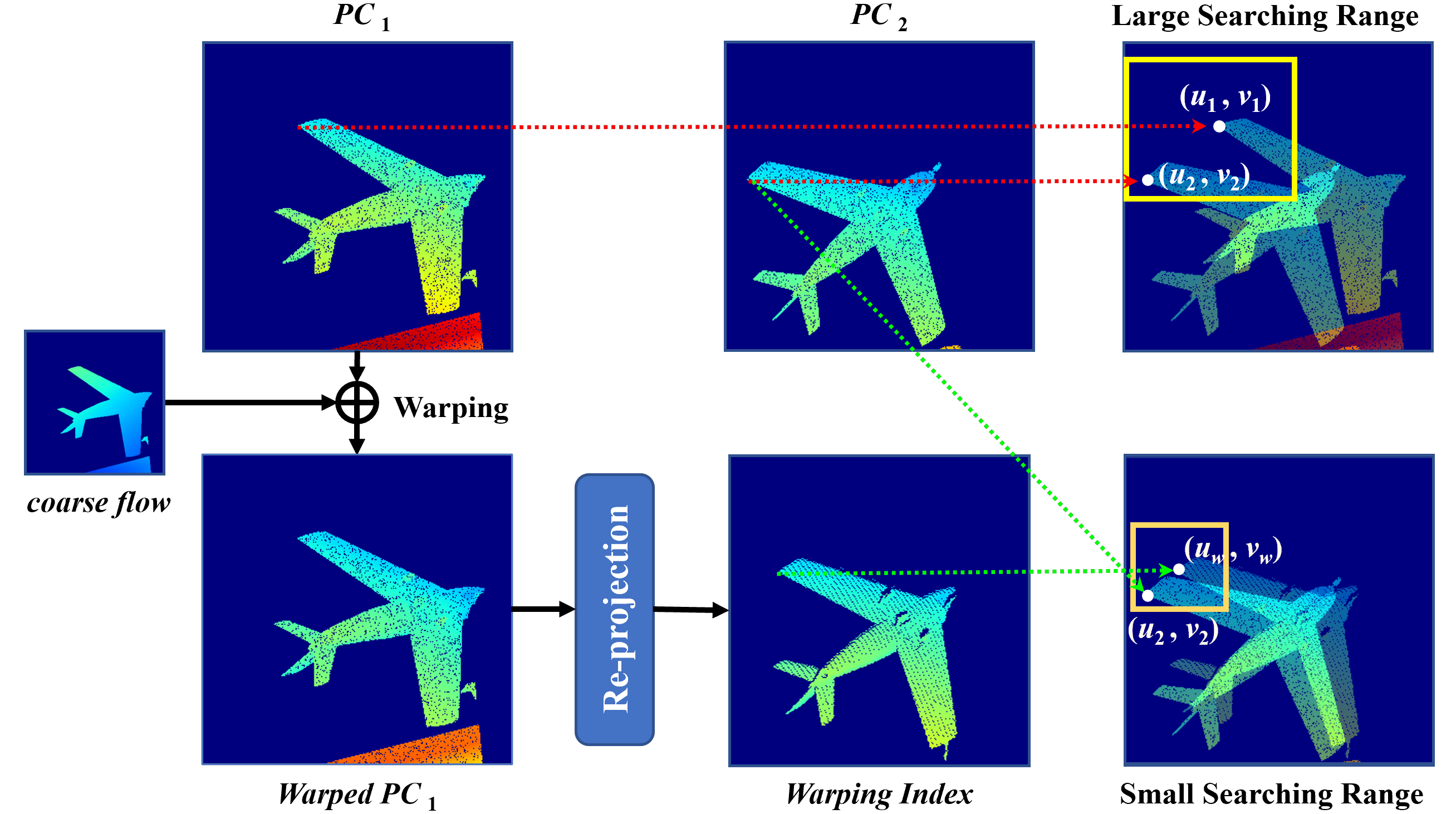}
    \vspace{-3mm}
	\caption{Cost volume computation with warping projection technique. The searching range can be reduced significantly. }
	\vspace{-3mm}
	\label{fig:warping}
\end{figure}

 In our setting, $p$ and $q$ are stored in a $h \times w \times 3$ matrix, located at their corresponding 2D grid $(u_1, v_1)$ and $(u_2, v_2)$ respectively. The warped points are obtained by adding the coarse flow $F_c \ (h \times w \times 3) $ to $PC_1$. In this way, $\hat{p} = p + f_{c}$ is stored in $(u_1, v_1)$ rather than their corresponding 2D grid $(u_w, v_w)$. We present a new cost volume module equipped with a warping projection technique to deal with this problem. As illustrated in Fig. \ref{fig:warping}, we create a $h \times w \times 2$ matrix to store the warping index $(u_{w}, v_{w})$ of every warped point in $PC_{1w}$.  
$(u_{w}, v_{w})$ acts as an indexing variable between $(u_1, v_1)$ and $(u_2, v_2)$ during the calculation of cost volume. For each warped point in $PC_{1w}$ located at $(u_1, v_1)$, a kernel is placed around the corresponding indexing position $(u_{w}, v_{w})$ on $PC_2$. Within the kernel, the distances between $PC_{1w}$ and $PC_2$ are calculated, and distant $PC_2$ points are filtered out while $K$ neighbors are retained.


During the computation of cost volume, each warped point has a corresponding 2D position due to our warping projection method. If we directly re-project $PC_{1w}$ to get $(u_w, v_w)$ and store them in the corresponding pixel, multiple points in $PC_{1w}$ could be mapped into the same grid and extra points would be merged. Our technique effectively avoids the loss of points resulted from merging operations. As a result, the complete point could be preserved. Our designed cost volume module also eliminates the need for a large kernel, thereby ensuring the processing efficiency.

\setlength{\tabcolsep}{1.5mm}
\begin{table*}[!t]
	\centering
	\footnotesize
	\begin{center}
 	{
    \begin{tabular}{ccccccc}
        \toprule
        \multirow{2}{*}{\text { Method }} & \multirow{2}{*}{\text { Input }} & \multicolumn{2}{c}{\text { 2D Metrics }} & \multicolumn{3}{c}{\text { 3D Metrics }}  \\
        & & $\text{EPE}_\text{2D}$ & $\text{ACC}_\text{1px}$ & $\text{EPE}_\text{3D}$ & $\text{ACC}_\text{0.05}$ & $\text{ACC}_\text{0.10}$ \\
        \midrule \text { FlowNet2.0 \cite{Ilg_2017_CVPR} } & \text { Image } & 5.05 & 72.8 \% & - & - & - \\
        \text { PWC-Net \cite{sun2018pwc} } & \text { Image } & 6.55 & 64.3 \% & - & - & - \\
        \text { RAFT \cite{teed2020raft} } & \text { Image } & 3.12 & 81.1 \% & - & - & - \\
        \midrule \text { FlowNet3D \cite{liu2019flownet3d}} & \text { LiDAR } & - & - & 0.169 & 25.4 \% & 57.9 \% \\
        \text { PointPWC-Net \cite{wu2019pointpwc} } & \text { LiDAR } & - & - & 0.132 & 44.3 \% & 67.4 \% \\
        \text { FLOT \cite{puy2020flot} } & \text { LiDAR } & - & - & 0.156 & 34.3 \% & 64.3 \% \\
        \midrule 
        \text { RAFT-3D \cite{teed2021raft} } & \text { Image+Depth } & 2.37 & 87.1 \% & 0.094 & 80.6 \% & -  \\
        \text { CamLiFlow \cite{liu2022camliflow}}  & \text { Image+LiDAR } & {2.20} & {87.3 \%} & {0.061} & 85.6 \% & 91.9 \%  \\
        \midrule
        \text { Ours } & \text { Image+LiDAR } & \bf{2.02} & {85.9 \%} & \bf{0.058} & \bf{86.7} \% & \bf{93.2 \%} \\
        \bottomrule
    \end{tabular}
    }
    \end{center}
    \vspace{-2mm}
    \caption{Quantitative results compared with recent methods on the FlyingThings3D dataset \cite{mayer2016large}. The performances are evaluated on all points (including occluded points and non-occluded points).}
    \vspace{-2mm}
  \label{table:ft3d}
\end{table*}

\begin{figure*}[!ht]
	\centering
	\includegraphics[width= 0.95\textwidth]{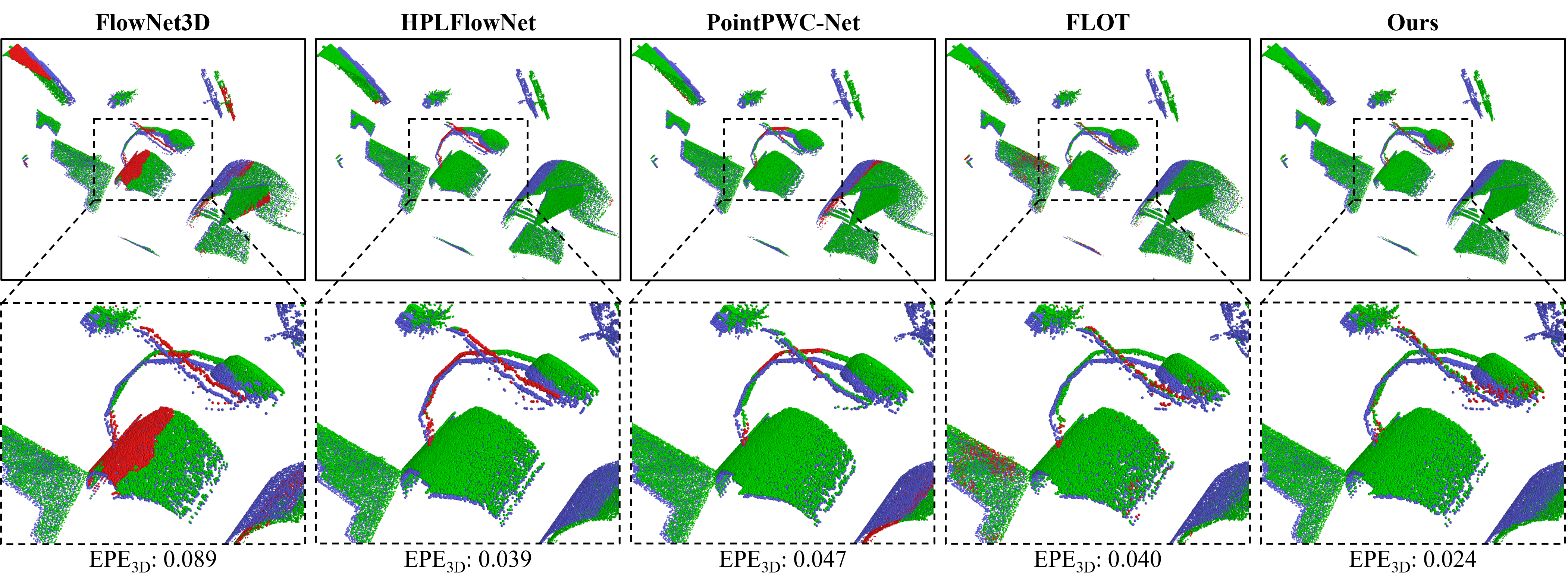}
	\vspace{-2mm}
	\caption{Qualitative visualization results of scene flow estimation. We compare our methods with FlowNet3D \cite{liu2019flownet3d}, HPLFlowNet \cite{gu2019hplflownet} and PointPWC-Net \cite{wu2019pointpwc} and FLOT \cite{puy2020flot}. Blue points represent $PC_1$. Green points represent accurate predictions and red points represent inaccurate predictions. (Accuracy is measured using $\text{ACC}_\text{0.10}$.)}
	\vspace{-4mm}
	\label{fig:vis}
\end{figure*}

\section{Experiments}

\begin{table*}[t]
	\centering
	\footnotesize
	\begin{center}
    {
    \begin{tabular}{lccccccccccccc}
        \toprule
        \multirow{2}{*}{\text { Method }} & \multirow{2}{*}{\text { Input }} & \multicolumn{3}{c}{\text { Disparity 1 }} & \multicolumn{3}{c}{\text { Disparity 2 }} & \multicolumn{3}{c}{\text { Optical Flow }} & \multicolumn{3}{c}{\text { Scene Flow }} \\
        & & \textit{bg} & \textit{fg} & all & \textit{bg} & \textit{fg} & all & \textit{bg} & \textit{fg} & all & \textit{bg} & \textit{fg} & all \\
        \midrule 
        SSF \cite{ren2017cascaded} & Stereo 
        & 3.55 & 8.75 & 4.42 
        & 4.94 & 17.48 & 7.02 
        & 5.63 & 14.71 & 7.14 
        & 7.18 & 24.58 & 10.07 \\
        Sense \cite{jiang2019sense} & Stereo 
        & 2.07 & 3.01 & 2.22 
        & 4.90 & 10.83 & 5.89 
        & 7.30 & 9.33 & 7.64 
        & 8.36 & 15.49 & 9.55 \\
        PRSM \cite{vogel20153d} & Stereo
        & 3.02 & 10.52 & 4.27 
        & 5.13 & 15.11 & 6.79 
        & 5.33 & 13.40 & 6.68 
        & 6.61 & 20.79 & 8.97 \\
        ACOSF \cite{li2021two} & Stereo
        & 2.79 & 7.56 & 3.58 
        & 3.82 & 12.74 & 5.31 
        & 4.56 & 12.00 & 5.79 
        & 5.61 & 19.38 & 7.90 \\
        DRISF \cite{ma2019deep} & Stereo
        & 2.16 & 4.49 & 2.55 
        & 2.90 & 9.73 & 4.04 
        & 3.59 & 10.40 & 4.73 
        & 4.39 & 15.94 & 6.31 \\
        M-FUSE \cite{mehl2023m} & Stereo
        & \textbf{1.40} & \textbf{2.91} & \textbf{1.65} 
        & 2.14 & 8.10 & 3.13
        & 2.66 & 7.47 & 3.46
        & 3.43 & 11.84 & 4.83 \\
        OpticalExp \cite{yang2020upgrading} & Mono + LiDAR
        & 1.48 & 3.46 & 1.81 
        & 3.39 & 8.54 & 4.25 
        & 5.83 & 8.66
        & 6.30 & 7.06 & 13.44 & 8.12 \\
        ISF \cite{behl2017bounding} & Stereo + LiDAR
        & 4.12 & 6.17 & 4.46 & 4.88 
        & 11.34 & 5.95 & 5.40 & 10.29 
        & 6.22 & 6.58 & 15.63 & 8.08 \\
        RigidMask+ISF \cite{yang2021learning} & Stereo + LiDAR
        &1.53	&3.65	&1.89
        &2.09   &8.92	&3.23	
        &2.63	&7.85	&3.50	
        &3.25	&13.08	&4.89	\\
        RAFT-3D \cite{teed2021raft} & Mono + Depth
        & 1.48 & 3.46 & 1.81 
        & 2.51 & 9.46 & 3.67 
        & 3.39 & 8.79 & {4.29}
        & {4.27} & 13.27 & {5.77} \\
        CamLiFlow \cite{liu2022camliflow} & Mono + LiDAR 
        & 1.48 & 3.46 & 1.81 
        & 1.92 & 8.14 & 2.95 
        & 2.31 & \textbf{7.04} & 3.10 
        & 2.87 & 12.23 & 4.43 \\
        \midrule
        Ours & Mono + LiDAR 
        & \textbf{1.40} & \textbf{2.91} & \textbf{1.65} 
        & \textbf{1.90} & \textbf{7.50} & \textbf{2.84} 
        & \textbf{2.27} & 7.10 & \textbf{3.07} 
        & \textbf{2.87} & \textbf{11.69} & \textbf{4.34} \\
        \bottomrule
    \end{tabular}
    }
    \end{center}
    \vspace{-2mm}
    \caption{Quantitative results compared with recent methods on KITTI Scene Flow dataset \cite{menze2018object} evaluated on all pixels (including occluded pixels). In the table, bg and fg denote the background and foreground, respectively. A lower value indicates better performance.}
  \label{table:kitti}
  \vspace{-2mm}
\end{table*}


\begin{figure*}[htpb]
	\centering
	\includegraphics[width= \textwidth]{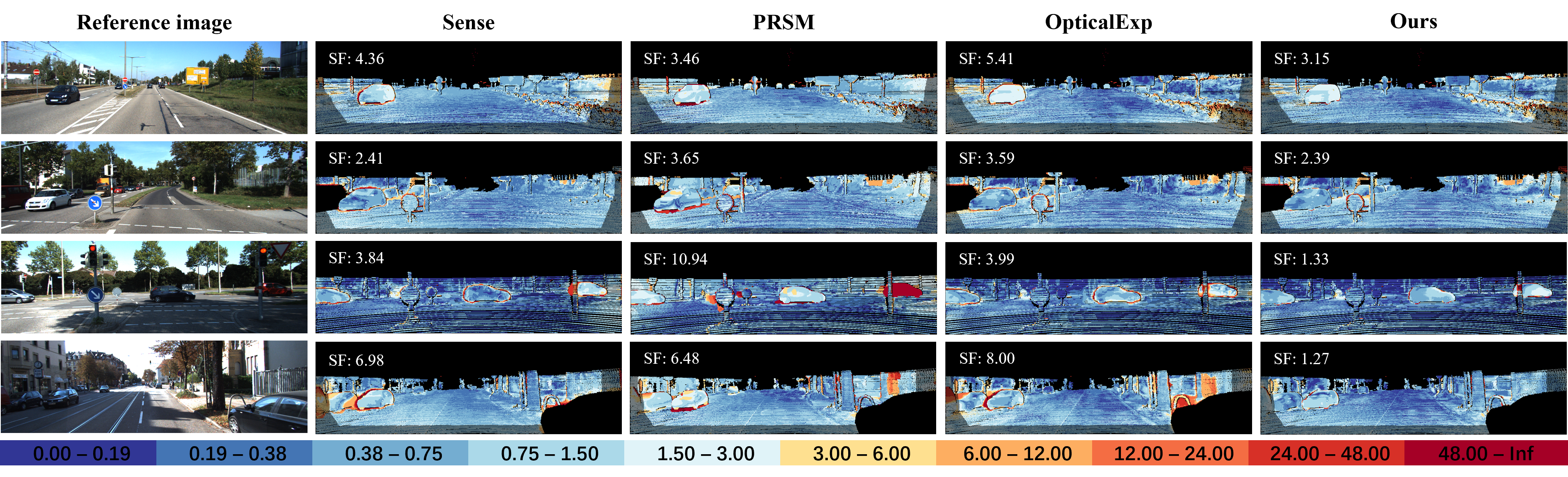}
	\vspace{-5mm}
	\caption{Qualitative visualization results of scene flow estimation on testing KITTI dataset. }
	\vspace{-4mm}
\label{fig:vis_kitti}
\end{figure*}

\subsection{FlyingThings3D}
\noindent{}{\bf Data Preprocessing:}
FlyingThings3D dataset is a synthetic dataset, consisting of 19640 training samples and 3824 testing samples. Following the preprocessing methods described in \cite{wu2019pointpwc, gu2019hplflownet, wang2021hierarchical}, the point clouds and ground truth scene flow are constructed from the depth map and optical flow.  We remove points with depth exceeding $35m$ as \cite{wu2019pointpwc, gu2019hplflownet, wang2021hierarchical}. We first train the model on a quarter of FlyingThings3D dataset to reduce training time, and then fine-tune the model on the complete dataset.

\vspace{5pt}
\noindent{}{\bf Evaluation Metrics:}
\label{sec:metrics_ft}
For a fair comparison, we adopt the following evaluation metrics as \cite{liu2019flownet3d, wu2019pointpwc, gu2019hplflownet, wang2022sfgan}: $\text{EPE}_\text{3D}$, $\text{ACC}_\text{.05}$, $\text{ACC}_\text{0.10}$ for 3D scene flow evaluation and $\text{EPE}_\text{2D}$, $\text{ACC}_\text{1px}$ for 2D optical flow evaluation. 

\vspace{5pt}
\noindent{}{\bf Quantitative results:}
The quantitative results on FlyingThings3D dataset are listed in Tab. \ref{table:ft3d}. We compare our method with other baselines \cite{Ilg_2017_CVPR, sun2018pwc, teed2020raft, liu2019flownet3d, wu2019pointpwc,puy2020flot,rishav2020deeplidarflow,teed2021raft, liu2022camliflow} in terms of the evaluation metrics described above. Results show that our method demonstrates comparable  performance both in 2D metrics and 3D metrics. 

\vspace{5pt}
\noindent{}{\bf Qualitative results:}
To demonstrate the effectiveness of our method for 3D scene flow prediction, we compare the performance of our model with other point-based methods \cite{liu2019flownet3d, gu2019hplflownet, wu2019pointpwc, puy2020flot} on FlyingThings3D dataset.
The qualitative results shown in Fig. \ref{fig:vis} show that our method achieves scene flow prediction with least errors.

\subsection{KITTI Scene Flow 2015}
\noindent{}{\bf Training:}
KITTI Scene Flow dataset contains 200 training samples and 200 testing samples.
Due to limited training samples, the model trained on FlyingThings3D are fine-tuned on KITTI dataset. Since KITTI dataset contains images with different size, we pad the images and the projected point clouds to a uniform size of $376 \times 1242$. 

\vspace{5pt}
\noindent{}{\bf Evaluation:}
The dense depth maps are obtained from LEAStereo \cite{cheng2020hierarchical} for testing. We reconstruct point clouds from the estimated depth with the calibration parameters. We submit our results to the official KITTI website, which adopts the percentage of outliers as evaluation metrics. 


\vspace{5pt}
\noindent{}{\bf Comparison with State-of-the-Arts:}
The quantitative results on KITTI Scene Flow dataset are listed in Tab. \ref{table:kitti}, which show that our method outperforms prior-arts. The visualization results are illustrated in Fig. \ref{fig:vis_kitti}.



\begin{table*}[t]
	\footnotesize
	\begin{center}
		{
			\begin{tabular}{l|l||cc|cccc}
				\toprule
				& &  \multicolumn{2}{c|}{\text {2D Metrics}}  & \multicolumn{4}{c}{\text { 3D Metrics }}  \\ 
				& \multirow{-2}{*}{\begin{tabular}[c]{@{}c@{}}Method \end{tabular}} &
				$\text{EPE}_\text{2D}$ & $\text{ACC}_\text{1px}$ & $\text{EPE}_\text{3D}$ & $\text{ACC}_\text{0.05}$ & $\text{ACC}_\text{0.10}$ & $\text{Outliers}$ \\
				\noalign{\smallskip}
				\hline\hline
				\noalign{\smallskip}
				(a)   
				& Ours (w/o projection, only 8192 input points)  
				& 2.23 & 82.2 \%
				& 0.061 & 84.3 \%	
				& 92.7 \% & 25.0 \% 
				\\
				&Ours (full, with projection, all points)     
				& \textbf{2.02} & \textbf{85.9 \%}
				& \textbf{0.058} & \textbf{86.7 \%}	
				& \textbf{93.2 \%}  & \textbf{20.8 \%}
				\\
				\noalign{\smallskip}
                \cline{1-8}\noalign{\smallskip}
				(b)   
				& Ours (w/o warping operation)  
				& 12.36 & 32.4 \%
				& 0.433 & 5.8 \%	
				& 20.0 \% & 97.6 \% 
				\\
				& Ours (w/o warping projection in cost volume)   
				& 4.53 & 74.9 \%
				& 0.105 & 49.9 \%	
				& 80.9 \% &57.5 \% 
				\\
				&Ours (full, with warping projection)     
				& \textbf{2.02} & \textbf{85.9 \%}
				& \textbf{0.058} & \textbf{86.7 \%}	
				& \textbf{93.2 \%}  & \textbf{20.8 \%}
				\\
				\noalign{\smallskip}
                \cline{1-8}\noalign{\smallskip}
				(c)   
				& Ours (w/o feature fusion)  
				& 4.32 & 76.8 \%
				& 0.081 & 69.5 \%	
				& 87.6 \% & 40.1 \% 
				\\
				& Ours (with concatenation feature fusion)   
				& 3.26 & 81.9 \%
				& 0.064 & 85.4 \%	
				& 92.6 \% & 21.5 \% 
				\\
				& Ours (full, with attentive feature fusion)   
				& \textbf{2.02} & \textbf{85.9 \%}
				& \textbf{0.058} & \textbf{86.7 \%}	
				& \textbf{93.2 \%}  & \textbf{20.8 \%}
				\\
				\noalign{\smallskip}
                \cline{1-8}\noalign{\smallskip}
				(d)   
				& Ours (w/o kernel-based grouping)  
				& {2.09} & 84.2 \%
				& \textbf{0.056} & {85.9 \%}	
				& \textbf{93.5 \%} & \textbf{19.9 \%} 
				\\
				& Ours (full, with kernel-based grouping)   
				& \textbf{2.02} & \textbf{85.9 \%}
				& {0.058} & \textbf{86.7 \%}	
				& {93.2 \%} & {20.8 \%}
				\\
				\bottomrule
			\end{tabular}
		}
	\end{center}
	\vspace{-2mm}
	\caption{The ablation study results of 3D scene flow learning on FlyingThings3D dataset \cite{mayer2016large}.}
	\vspace{-3mm}			
	\label{table:ablation}
\end{table*}

\subsection{Implementation Details}
All experiments are conducted on NVIDIA Quadro RTX 8000 GPU with PyTorch 1.10.0. We use Adam \cite{adam} as our optimizer with $\beta_1 = 0.9, \beta_2 = 0.999$. The initial learning rate of 0.001 decays exponentially with a decaying rate $\gamma = 0.8$, and the decaying step is 80. Our designed network adopts the hierarchical structure, where both low-level and high-level flow are predicted. The optical flow and scene flow are predicted similarly, both in a pyramid, warping, and cost volume (PWC) structure as Fig. \ref{fig:sto}. Therefore, we adopt a joint multi-scale supervised manner to train the model by adding optical flow loss and scene flow loss. Let $F^l = \{ {f_{i}^{l}}|{f_{i}^{l}} \in {\mathbb{R}^3}\} _{i = 1}^{N_l}$ be the predicted flow at level $l$, and $\hat{F^l} = \{ {\hat{f_{i}^{l}}}|{\hat{f_{i}^{l}}} \in {\mathbb{R}^3}\} _{i = 1}^{N_l}$ be the corresponding ground truth flow. The training loss of optical flow and scene flow is calculated in the same way as follows:
\begin{equation}
loss=\sum_{l=1}^{4} w_{l} \frac{1}{N_{l}} \sum_{i=1}^{N_{l}}\left\|\hat{f_{i}^{l}}-f_{i}^{l}\right\|_{2},
\end{equation}
where $N_l$ denotes the number of valid points or pixels at level $l$. $w_l$ denotes the weighting factor of losses at different levels, and $w_1 = 0.1, w_2 = 0.2, w_3 = 0.3, w_4 = 0.8$. 


\subsection{Ablation Study}
We conduct a series of ablation studies to verify the effectiveness of each component of our proposed network on the validation set of FlyingThings3D dataset. 

\vspace{5pt}
\noindent{}{\bf Dense or Sparse Representation:}
We compare the performances of our method with or without 2D dense representation of point clouds. For experiments without the 2D dense representation, following \cite{wu2019pointpwc, wang2021hierarchical}, we randomly select 8192 points and take them as input for the neural network. As 2D representation allows us to input more information, the model using dense representation outperforms the sparse 3D representation as shown in Tab. \ref{table:ablation} (a).

\vspace{5pt}
\noindent{}{\bf Warping Projection in Cost Volume:}
For experiments without our proposed warping projection technique, we project directly the $PC_{1w}$ onto 2D plane and merge multiple projected points in the same grid. The flow embedding features for merged points are obtained from interpolation. Results in Tab. \ref{table:ablation} (b) demonstrate that the  warping projection technique is useful in avoiding information loss and thus improve the model's performance.

\vspace{5pt}
\noindent{}{\bf Attentive Feature Fusion:}
For experiments without feature fusion, the network takes only point clouds as input. Non-attentive feature fusion simply concatenates the image features and point cloud features. Our feature fusion approach utilizes an attention mechanism to aggregate features in a pixel-point manner. We find that feature fusion improves prediction accuracy, and the attention-based fusion reports the best performance, as shown in Tab. \ref{table:ablation} (c).

\vspace{5pt}
\noindent{}{\bf Kernel Based Grouping:}
For experiments without kernel based grouping, the network use KNN algorithm to group neighbors among all points. As shown in Tab. \ref{table:ablation} (d), similar or even better performance is obtained because the searching range includes all points. However, the searching process becomes computationally expensive. Therefore, we compare the latency of our model with and without the kernel based grouping technique. The results in Tab. \ref{table:comparison_of_time} show that the efficiency drops by a factor of 20.

\subsection{Analysis}

The experimental results demonstrate that our proposed method achieves comparable performance on both FlyingThings3D and KITTI dataset. We attribute this improvement to three key components of our approach: First, our approach benefits from the dense representation. Previous methods often take a sparse set of points as input, typically 8192 points, which is much less than the total points in a scene. This downsampling operation leads to a loss of details and may result in an incomplete representation of the scene. In contrast, we store unordered 3D points in regular 2D grids. The dense representation provides a more precise structure of the scene, allowing us to preserve complete information without losing points. 
 Second, the proposed cost volume with the warping projection technique significantly reduces the size of the search kernel, thus improving the computational efficiency. Unlike previous warping methods that directly project the warped points onto 2D plane and merge extra points projected into the same grid, our new warping projection technique enables us to avoid such information loss by storing the corresponding warped 2D coordinates as intermediate indexes. 
  Third, the multi-modal feature fusion further improves the motion prediction. The integration of image and point cloud is advantageous as they are complementary to each other. Images contain rich color information, while point clouds provide abundant 3D geometrical information. Our attentive feature fusion between these modalities contributes to a more precise and distinctive representation of points. Overall, our approach leverages more input data and reduces information loss to improve the prediction accuracy of point-wise motion.

\setlength{\tabcolsep}{1.3mm}
\begin{table}[t]
	\centering
	\footnotesize
	\begin{center}
	{
      \begin{tabular}{lccc}
        \toprule 
        Methods & Input point number & Size (MB) & Time (ms)\\
        \midrule
        HPLFlowNet \cite{gu2019hplflownet} & 8192 &  231.8 & 93.67 \\
        HPLFlowNet \cite{gu2019hplflownet} & 49152 & 231.8 & 132.58 \\
        PointPWC-Net \cite{wu2019pointpwc} & 8192 & 30.1 & 100.05 \\
        PointPWC-Net \cite{wu2019pointpwc} & 49152 & 30.1 & 2643.61 \\
        Ours (w/o KBG) & 56269 (480 $\times$ 270) & \textbf{20.1} & 1324.29 \\ 
        Ours & 56269 (480 $\times$ 270) & \textbf{20.1} & \textbf{52.91} \\ 
        \bottomrule
      \end{tabular}
     }
    \end{center}
    \vspace{-2mm}
    \caption{Comparison of model size and running time. KBG denotes the Kernel Based Grouping technique.}
    \label{table:comparison_of_time}
    \vspace{-5mm}
\end{table}


\vspace{5pt}
\noindent{}{\bf Latency:} We further evaluate the performance of our model by comparing the running time and size with other methods. As shown in Tab. \ref{table:comparison_of_time}, our framework outperforms others in terms of efficiency. Since the number of valid points may vary across frames, we calculate the average number of valid points in the validation split of the dataset.

\vspace{5pt}
\noindent{}{\bf Limitations:} The projection-based framework requires the input point cloud to be 2.5D format, otherwise multiple points could be mapped into the same grid. In the future, we plan to introduce hash map to store multiple points to address this issue. Currently, our method is applicable to FlyingThings3D and KITTI dataset since the point clouds are converted from depth map. Considering the surface-scanning characteristic of LiDAR sensor, our method is still practical in autonomous driving. 

\section{Conclusion}
In this paper, we propose DELFlow, allowing us to take the entire point clouds of a scene as input and efficiently estimate scene flow at once. To preserve the complete information of raw point clouds without losing details, we introduce a cost volume module based on warping projection. Besides, the attentive feature fusion from images and point clouds further improves the accuracy. By leveraging more data input and reducing information loss, our approach outperforms prior-arts in efficiency and effectiveness.

{\bf \small  Acknowledgement.} This work was supported in part by the Natural Science Foundation of China under Grant 62225309, 62073222, U21A20480 and U1913204.

\newpage
{\small
\bibliographystyle{ieee_fullname}
\bibliography{egbib}
}

\end{document}